\documentclass[10pt,twocolumn,letterpaper]{article}

\usepackage{cvpr}
\usepackage{times}
\usepackage{epsfig}
\usepackage{graphicx}
\usepackage{amsmath}
\usepackage{amssymb}
\usepackage{lipsum}
\usepackage{caption}
\usepackage{mwe}

\cvprfinalcopy 

\ifcvprfinal\fi
\title{SynthCam3D: Semantic Understanding With Synthetic Indoor Scenes}

\author{
\begin{tabular}[t]{c@{\extracolsep{1em}}c@{\extracolsep{1em}}c@{\extracolsep{1em
}}c@{\extracolsep{1em}}c} 
Ankur Handa & Viorica P{\u a}tr{\u a}ucean  & Vijay Badrinarayanan & Simon Stent 
& Roberto Cipolla \\
{\tt\small ah781@cam.ac.uk} & {\tt\small vp344@cam.ac.uk} & {\tt\small 
vb292@cam.ac.uk} & {\tt\small sais2@cam.ac.uk} & {\tt\small 
rc10001@cam.ac.uk} \\
& & \\
 \multicolumn{5}{c}{Department of Engineering, University of Cambridge, UK}
\end{tabular}
}

\begin{document}
\twocolumn[{%
\renewcommand\twocolumn[1][]{#1}%
\maketitle
\begin{center}
    \centering
    \includegraphics[width = 1.05\linewidth]{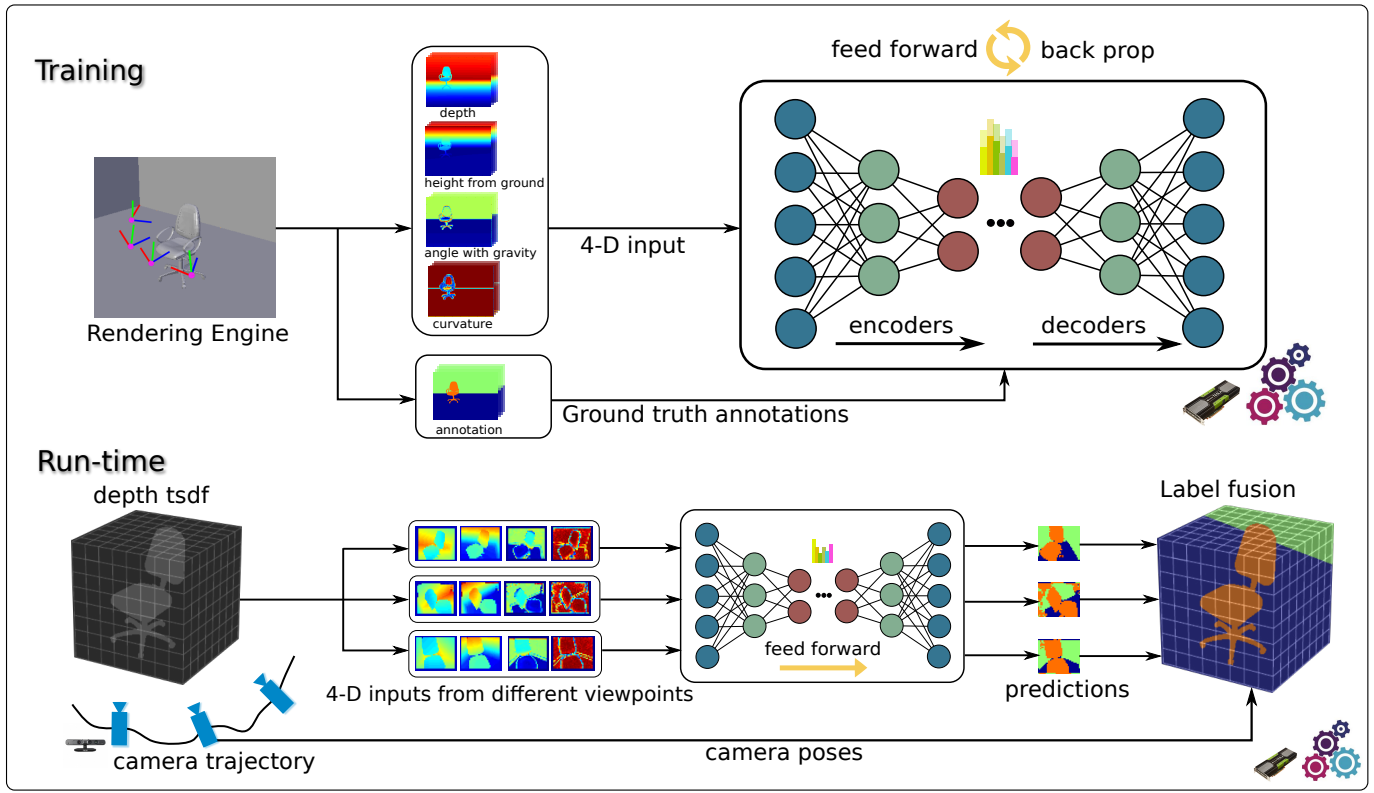}
    \captionof{figure}{Our system is trained exclusively on synthetic data obtained from our scene library, SynthCam3D. During testing, per-frame predictions returned by the network are fused using the camera poses provided by the reconstruction system.}
 \label{fig:synthcam3d}
\end{center}%
}]



\begin{abstract}
We are interested in automatic scene understanding from geometric cues. To this end, we aim to bring semantic segmentation in the loop of real-time reconstruction. Our semantic segmentation is built on a deep autoencoder stack trained exclusively on synthetic depth data generated from our novel 
3D scene library, \emph{SynthCam3D}. Importantly, our network is able to segment real world scenes without any noise modelling. We present encouraging preliminary results. 

   
\end{abstract}


\section{Introduction}
\begin{figure*}
 \includegraphics[width = \linewidth]{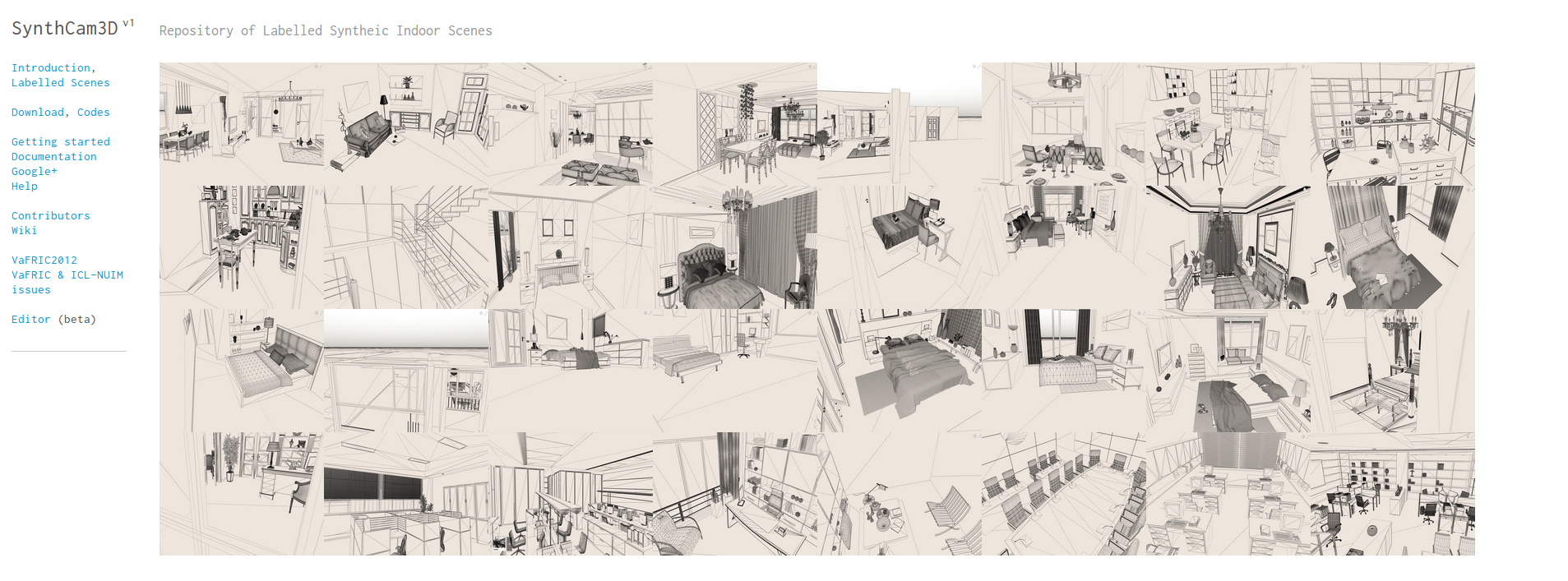}
 \caption{SynthCam3D is a library 
of synthetic indoor 
scenes collected from various online 3D repositories and hosted at  
\texttt{http://robotvault.bitbucket.org.}}
 \label{fig:synthcam3d}
\end{figure*}
Fully automatic understanding of 3D scenes is of particular interest for many
attractive applications that demand interaction with objects and/or 
primitive parts that make up the scene \cite{farabet2013,Silberman:ECCV12}. Such knowledge is indispensable for a robot to be able to perform fully autonomously basic interactions with its environment, like moving objects, clearing the clutter, stacking objects on top of others, or searching for objects in their likely locations. These actions require richer understanding of the scene than \eg the per-image labels from image classification approaches or object bounding boxes provided by object detectors. 

We believe that a key step towards whole scene understanding is the semantic segmentation of the scene. Our 
work brings together two established directions towards 
the goal of 3D scene understanding: 3D reconstruction and deep learning-based 
semantic 
segmentation. Here, we exploit the inherent dependency between reconstruction 
and segmentation --- per-frame labels are fused using their respective camera 
poses returned by the reconstruction system. In doing so, we particularly stress 
the importance of treating data coming as a video stream. 
On an average, segmentations from different viewpoints, when 
fused, should yield a result better than a segmentation from any particular 
view. Our system is directly related to Hermans \etal's 
work \cite{Hermans14ICRA}, who fuse per-frame segmentations 
obtained with randomised decision forests from RGB-D images; they use 2D and 3D 
dense CRFs \cite{denseCRF} to smooth the per-frame 2D segmentations and the 
fused 3D segmentation, respectively. We harness recent advances made in deep learning to 
obtain per-frame dense predictions. Our deep architecture, inspired from 
\cite{Ranzato:2007}, is composed of stacked autoencoders and trained 
modularly. For all our experiments, we use depth data as the only cue for 3D 
scene understanding. The motivation of using depth images is twofold: firstly, 
depth discontinuities are very important for object recognition as has been 
shown in \cite{guptaECCV14}, and secondly, the convenience in obtaining depth 
data. Using only depth cues spares us from the complications of dealing 
with the infinite space of possible textures and lighting setups, making it 
tractable to collect a representative set of scenes in terms of scene layout and 
objects distribution. The challenge in this context is to investigate if depth 
data is a \textit{sufficient} input for semantic segmentation.



We make publicly available a new library -- SynthCam3D -- 
consisting of a significant number of labelled synthetic 3D scenes and 
associated code for generating depth maps and their corresponding annotations.  The 
scenes belong to different semantic categories and have been compiled together from various online 3D repositories 
\cite{crazy3dfree}, and manually annotated. Large public repositories (\eg 
Trimble Warehouse) of 3D CAD models have existed in the past, but they have 
mainly served the graphics community. It is only recently that we have started 
to see emerging interest in synthetic data for computer vision. The advantages 
of synthetic 3D models cannot be overstated, especially when considering 
scenes: once a 3D annotated model is available, it allows rendering as many 2D 
annotated views as desired, at any resolution and frame-rate. In comparison, 
existing datasets of real data are fairly limited both in the number of 
annotations and the amount of data. NYUv2 \cite{Silberman:ECCV12} provides only 
795 training images for 894 classes; hence 
learning any meaningful features characterising a class of objects becomes 
prohibitively hard. SynthCam3D is particularly useful for:

\begin{figure*}[htp]
\centerline{
\includegraphics[width=0.125\linewidth]{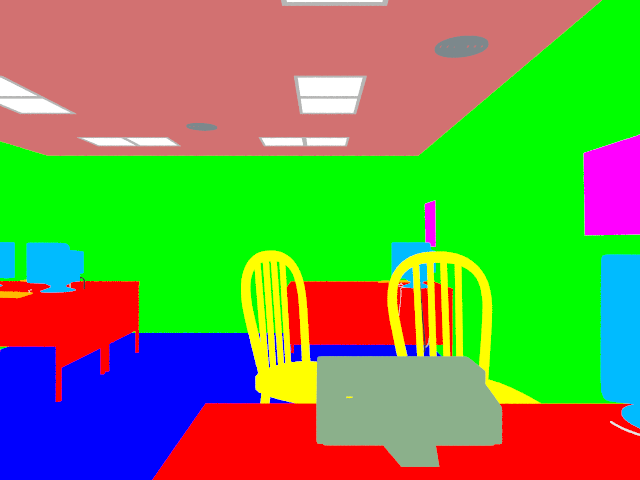}
\includegraphics[width=0.125\linewidth]{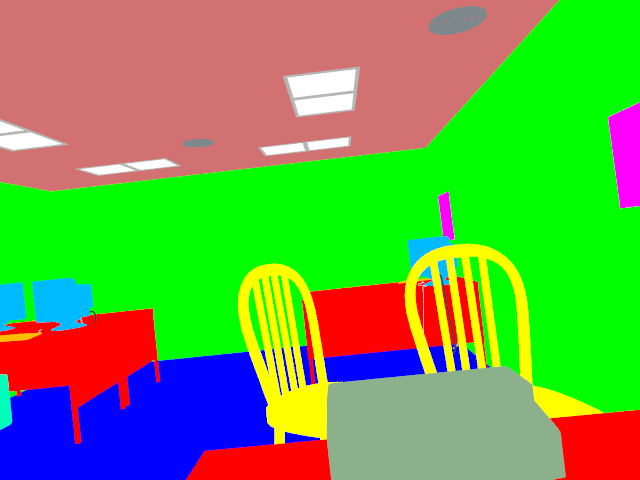}
\includegraphics[width=0.125\linewidth]{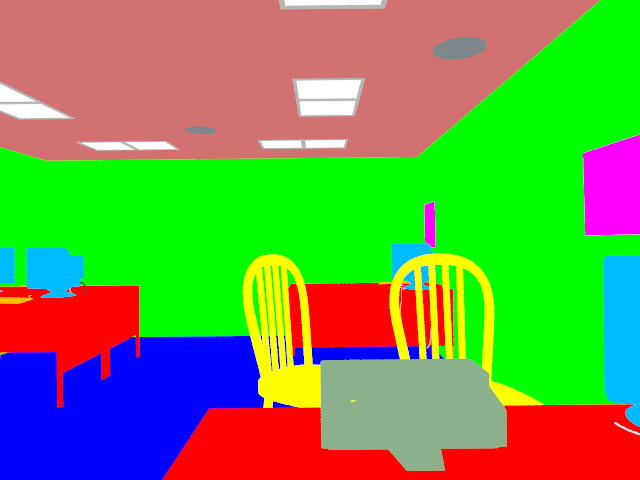}
\includegraphics[width=0.125\linewidth]{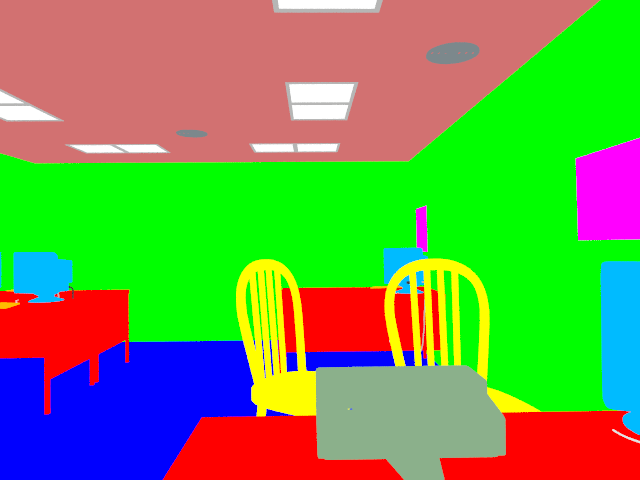}
\includegraphics[width=0.125\linewidth]{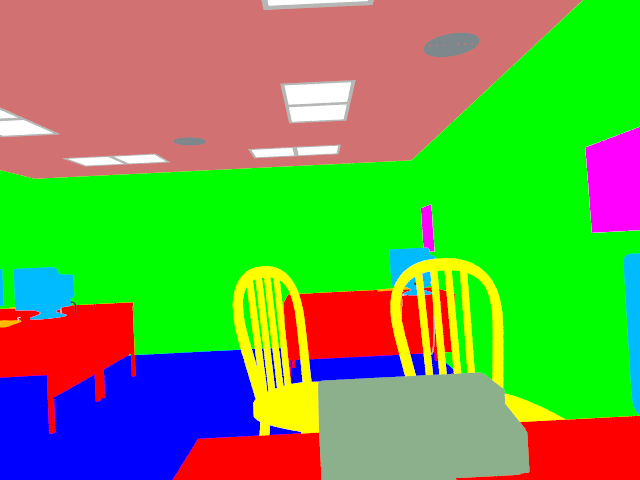}
\includegraphics[width=0.125\linewidth]{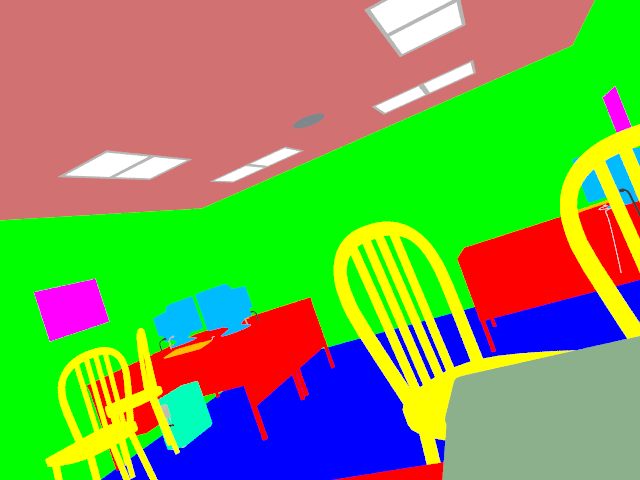}
\includegraphics[width=0.125\linewidth]{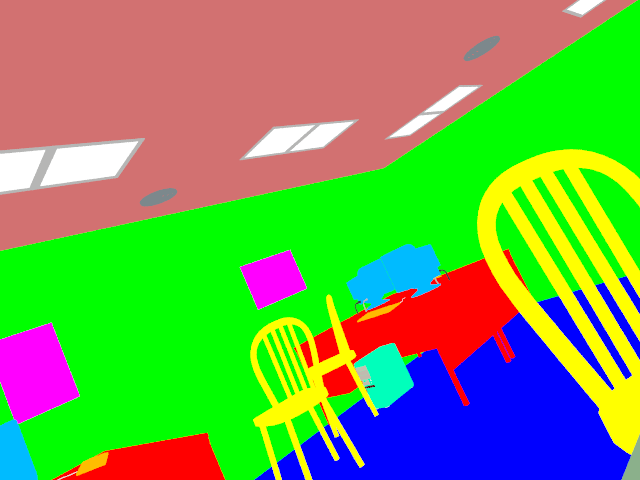}
\includegraphics[width=0.125\linewidth]{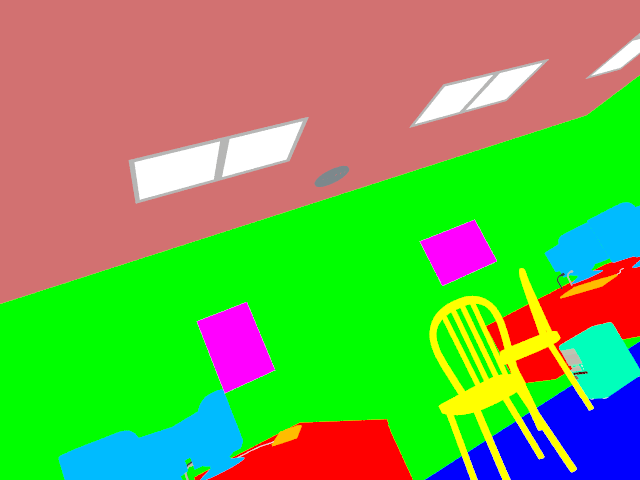}
}
\caption{\small
Samples of annotated images rendered at various camera poses for an office 
scene taken from SynthCam3D.} 
\label{fig:annots}
\end{figure*}

\begin{itemize}
 \item Generating potentially unlimited high-quality annotated depth data for different types 
of scenes (Fig. \ref{fig:annots}). 
 \item Benchmarking large scale depth-only SLAM systems on complex scenes, by providing 
ground truth geometry \cite{HandaECCV2012} \cite{handa:ICRA2014}. 
 \item Enabling training generative models similar to \eg \cite{liu14sigasia}, 
to learn common scene layouts and object relationships, which can then be used 
to synthesize more scenes effortlessly.

\end{itemize}

In the following, we describe SynthCam3D and briefly outline our system 
trained using data generated from the library. Preliminary results show the 
usefulness of the proposed library for training deep architectures for semantic 
segmentation of real world scenes. With a careful choice of input features to 
our deep learning network and using depth maps raycasted by the 
reconstruction system, we are able to bypass the domain adaptation issues 
that have been observed in the past \ie the system trained on synthetic depth data can be directly applied to segment real depth data, without the need of noise modelling at training time.

\section{SynthCam3D Library}
\begin{table}[h]
\centering
\begin{tabular}{|l|c|}
\hline
\textbf{Category}      & \textbf{Number of 3D models} \\ \hline
Bedrooms      & 11                  \\ \hline
Office Scenes & 15                  \\ \hline
Kitchens      & 11                  \\ \hline
Living Rooms  & 10                  \\ \hline
Bathrooms     & 10                  \\ \hline
\end{tabular}
\caption{\small Different scene categories and the number of annotated 3D models for each category.}
\end{table}
SynthCam3D contains 3D models from five different scene categories: bedroom, 
office, kitchen, living-room, and bathroom, with at least 10 annotated scenes 
per category. Importantly, all the 3D models are in metric scale. Each scene is 
composed of up to around 50--150 objects and the complexity can be controlled 
algorithmically. The granularity of the annotations can be adapted by the user 
depending on the application, \eg in our experiments on bedroom scenes we 
condensed the number of classes down to 15 for generating data and understanding 
only functional categories of objects. The models are provided in \textit{.obj} 
format, together with the code and camera settings needed to set up the 
rendering using POV-Ray. A simple OpenGL based GUI allows the user to place 
virtual cameras in the synthetic scene at desired locations to generate a 
possible trajectory for rendering at different viewpoints. Fig. 
\ref{fig:annots} shows samples of rendered annotated views of a simple office 
scene.       


\section{Rendering Engine}
We use the popular ray-tracer POVRay for our rendering purposes, being inspired 
by the past work of Handa \etal \cite{HandaECCV2012}. To render depth maps with 
associated annotations from the \textit{.obj} 
models, we first need to convert the \textit{.obj} models to their 
corresponding POVRay files using 
\textsf{Poseray}\footnote{\url{https://sites.google.com/site/poseray/}.}. Then 
the camera extrinsic parameters are set with a 3$\times$4 matrix inside the 
main POVRay file (having the \textit{.pov} extension). Eventually, a rendering 
trajectory can be obtained by varying the camera parameters inside the main 
POVRay file. Each rendering operation outputs an annotation file, a depth map, 
and a text file containing the associated camera intrinsic and extrinsic 
parameters. These files are parsed with the codes available from 
\cite{handa:ICRA2014}. Since we only need depth and annotations, the rendering 
procedure is fast, taking less than one second per view on a standard desktop 
machine for VGA resolution.
%

\section{System Overview}

Our system relies on reconstruction front-end running in real-time and deep 
learning back-end that takes in 4D input channels namely, depth, height from 
ground, angle with gravity vector, and curvature (DHAC). The labels obtained 
from different viewpoints are then fused together with the classic Bayesian 
filtering \cite{Thrun:2005} on a voxelised volume using the camera poses 
returned by the reconstruction system. We observe immediate benefits of 
performing 3D mapping and semantic segmentation in parallel threads: first, at 
test time, we can use depth maps raycasted from the mapping volume, which have 
superior quality compared to raw depth maps; this results in improved 
segmentation results. Second, we can 
improve the overall segmentation of the scene by label fusion. We briefly describe reconstruction and our deep learning architecture below.

\subsection{Reconstruction} 
Our reconstruction system is a custom implementation of the well-known KinectFusion  algorithm \cite{Newcombe:2011}, wherein depth maps are averaged with their truncated sign distance representation on a voxelised 3D volume. For all our segmentation experiments, we use raycasted depth maps and camera poses obtained via this system module. Finally, we align the local reference frame of the reconstruction with the 
inertial frame, using the simple and effective optimisation proposed in 
\cite{Gupta:CVPR13} to obtain the required rotation matrix. This allows us to 
compute features that are invariant to rotation about the gravity axis, \ie 
height from the ground plane and angle with gravity vector. 




\subsection{Segmentation using deep learning} 
Our segmentation module is inspired by the deep architecture used 
in \cite{Ranzato:2007}. It is composed of a sequence of stacked auto-encoders, 
with supervised modular training of each layer to capture the representative 
features of the scene at different scales and produce dense predictions for 
each pixel in the input depth map. We use this architecture primarily due to its 
lightweight structure, compared to \eg \cite{long_shelhamer_fcn}, which has 
prohibitive memory requirements. 

\begin{figure*}
\centerline{
\includegraphics[width=0.5\linewidth]{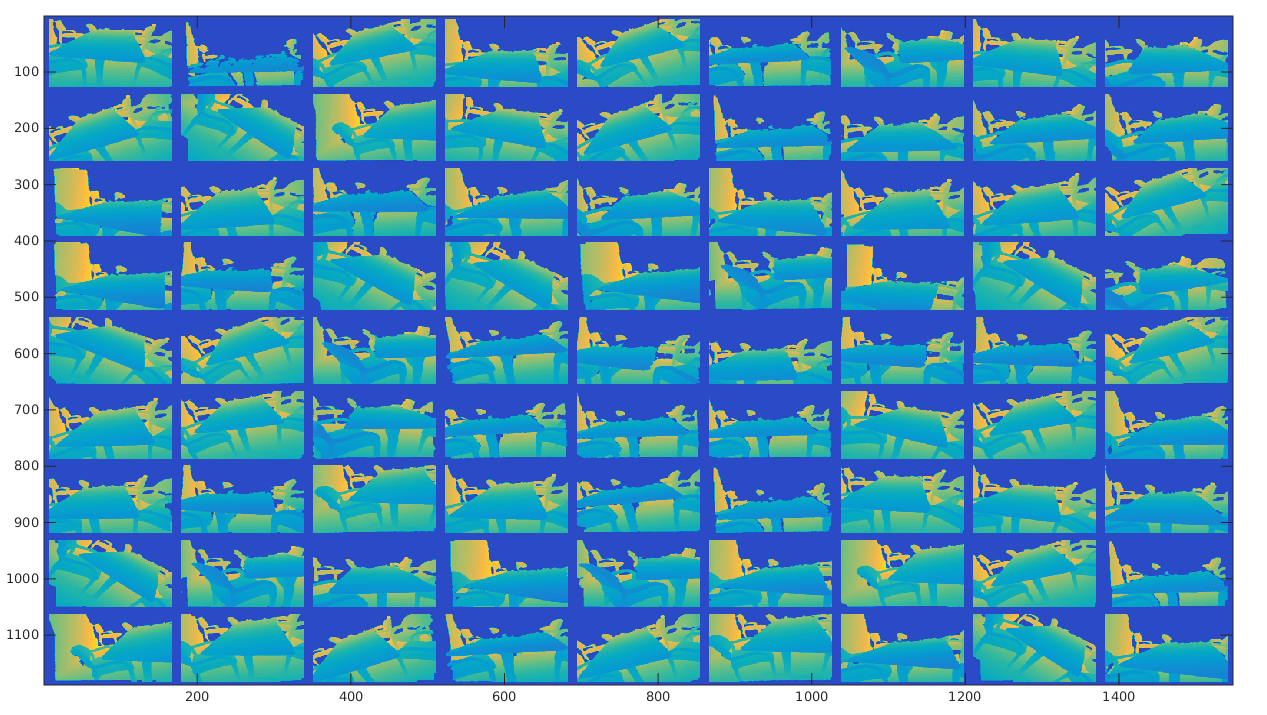}
\includegraphics[width=0.496\linewidth]{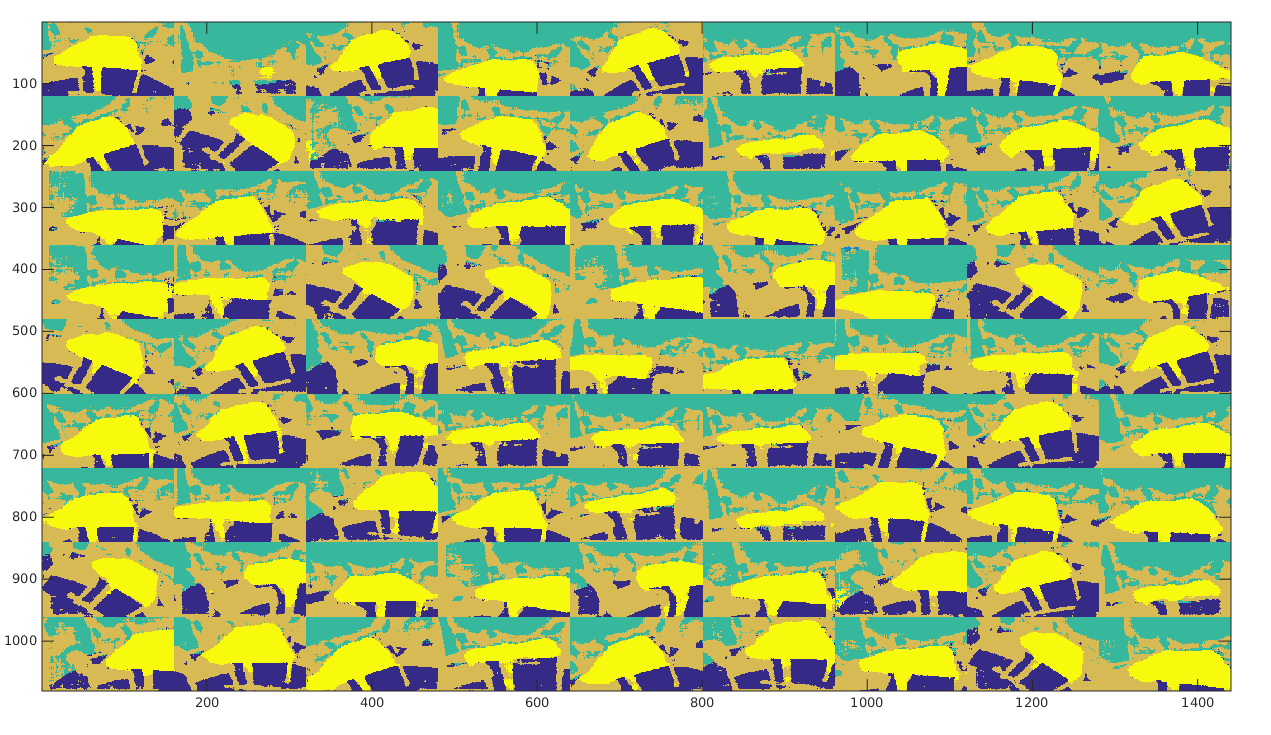}
}
\centerline{
\includegraphics[width=0.5\linewidth]{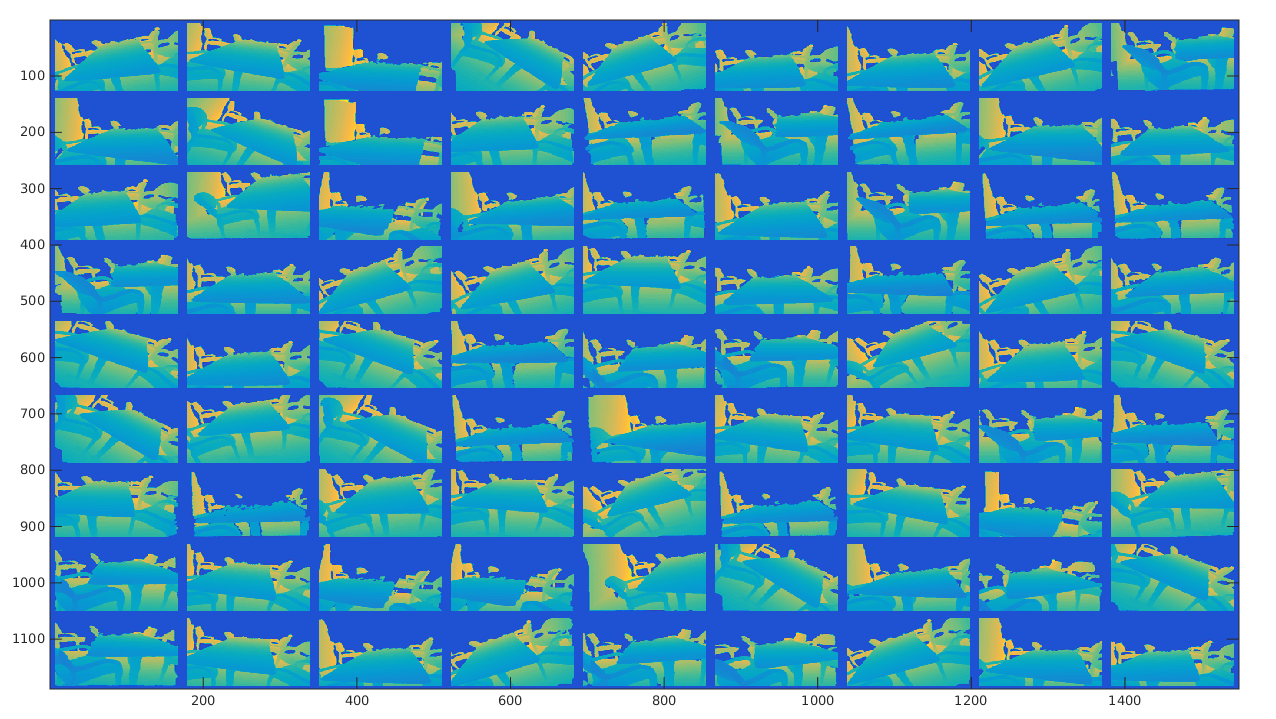}
\includegraphics[width=0.5\linewidth]{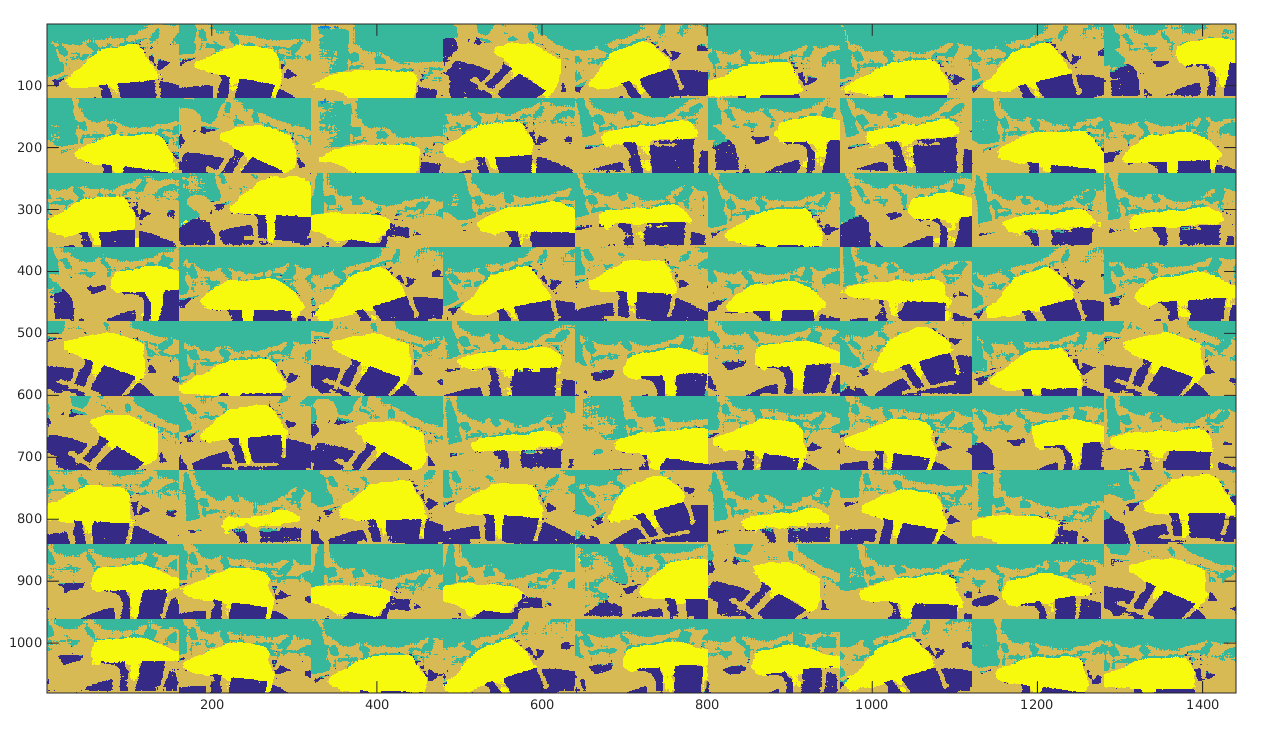}
}
\caption{\normalsize
Real data results on tables and chairs. First column shows the depth images raycasted from the tsdf volume and second column shows the segmentation results.}
\label{fig:results_on_chairs_tables}
\end{figure*}

\begin{figure*}
\vspace{2mm}
\centerline{
\includegraphics[width=0.5\linewidth]{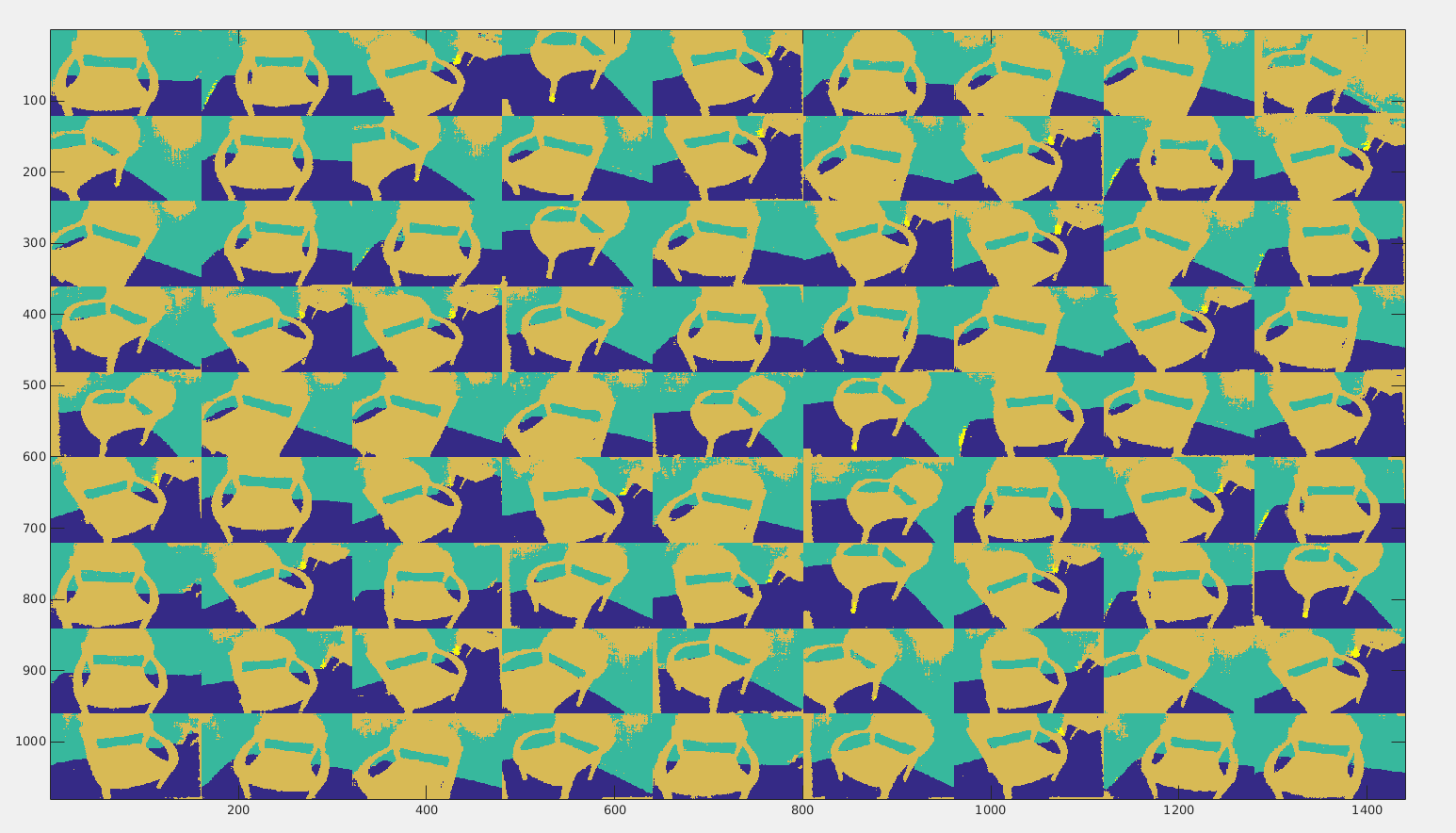}
\includegraphics[width=0.495\linewidth]{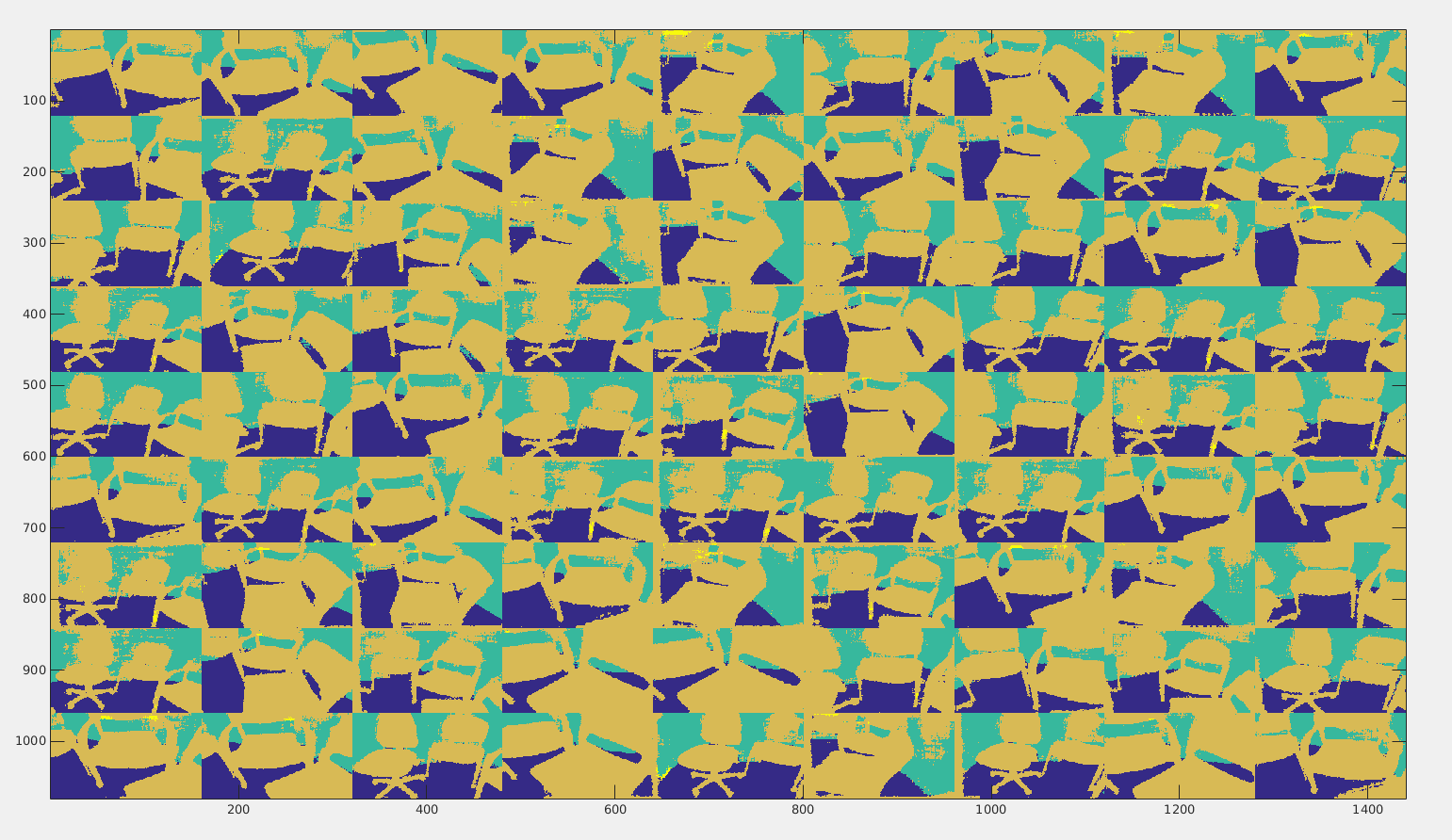}
}
\caption{\normalsize
Left: results on one chair. Right: results 
on multiple chairs. Note that the training was done on scenes containing both 
chairs and tables. }
%
\label{fig:chairs}
\end{figure*}

We perform preliminary experiments with this network on simple scenes composed 
of chairs and tables. In all our experiments, we segment the scene into 5 different classes: chairs, tables, floor, ceiling, and wall. Figure \ref{fig:layers} shows the segmentation results on training data where a clear improvement of the results is evident as layers are added progressively to the network. Figure \ref{fig:results_on_chairs_tables} and \ref{fig:chairs} show results on real world scenes where we are able to get good segmentations; the training was done exclusively on synthetic scenes containing chairs and tables.

Video Links: \url{http://robotvault.bitbucket.org/results.html}
\begin{figure*}
\centerline{
\includegraphics[width=0.25\linewidth]{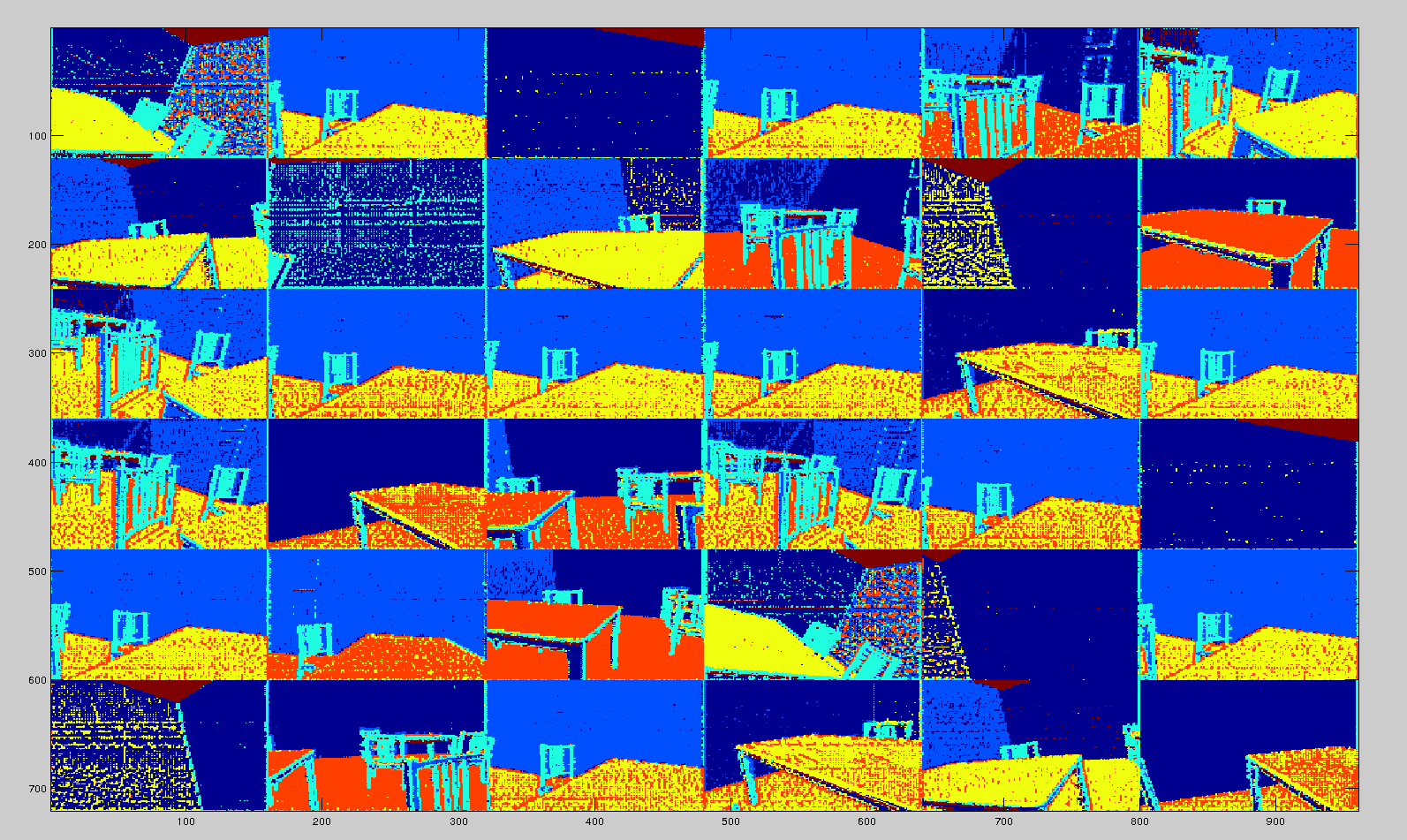}
\includegraphics[width=0.25\linewidth]{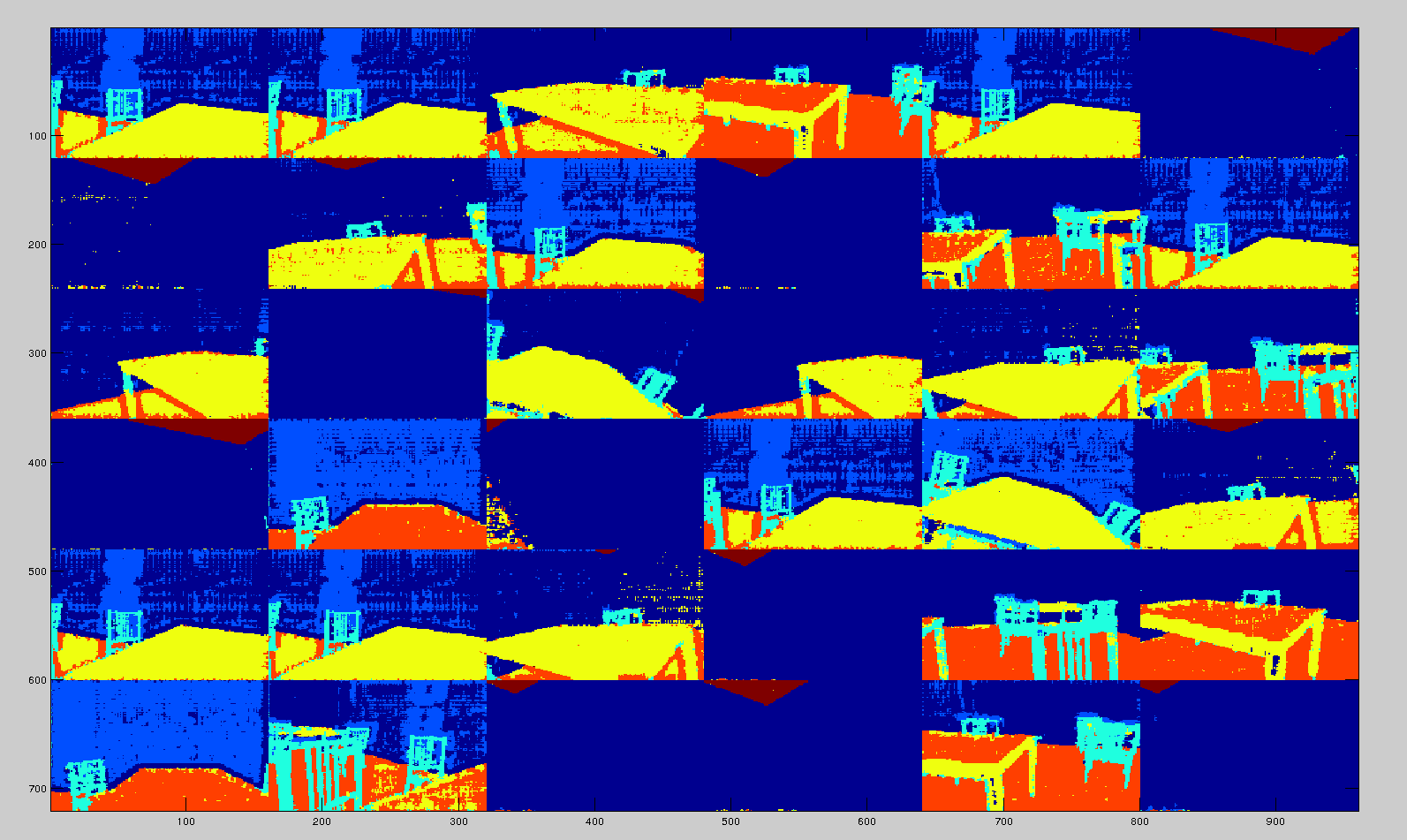}
\includegraphics[width=0.25\linewidth]{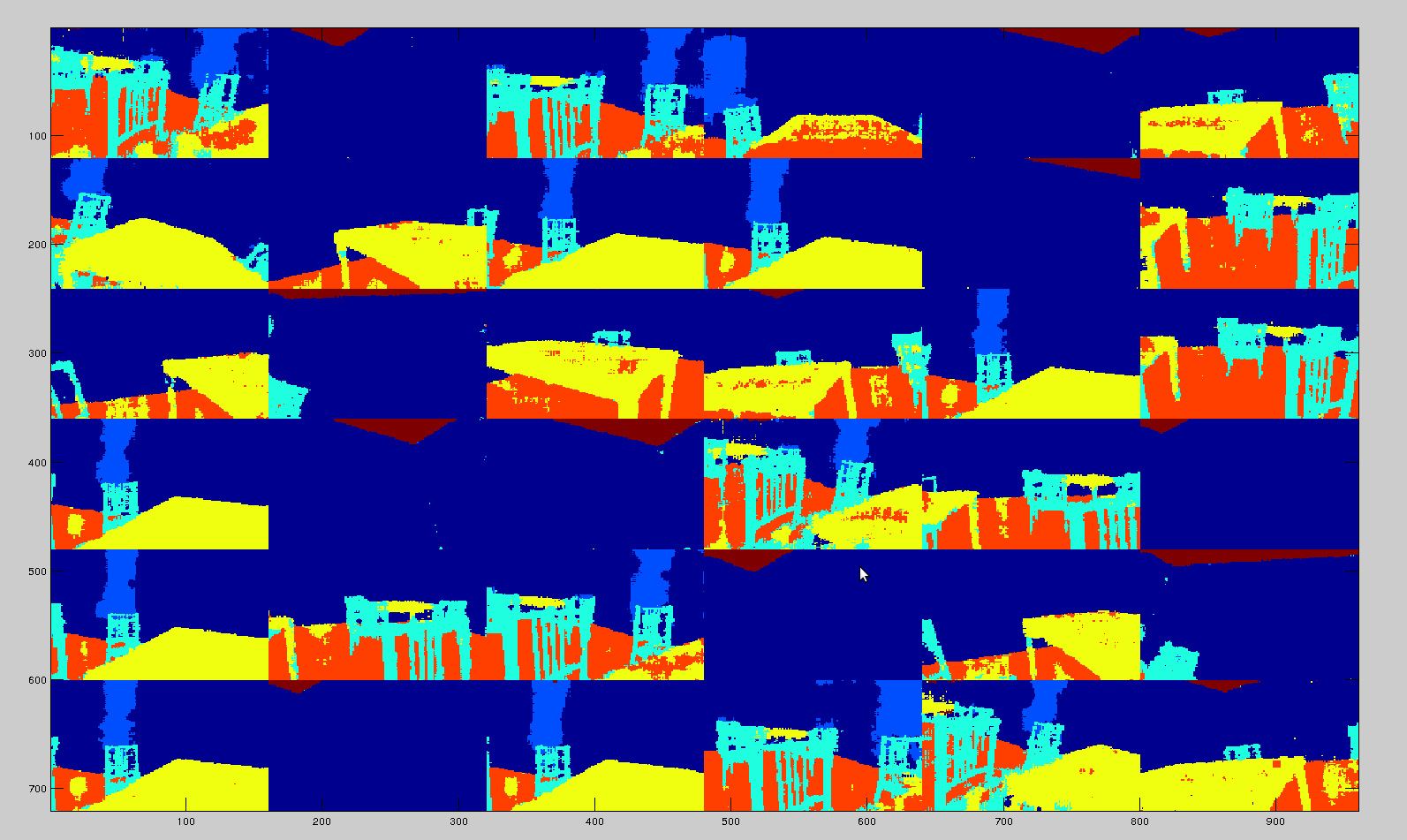}
\includegraphics[width=0.25\linewidth]{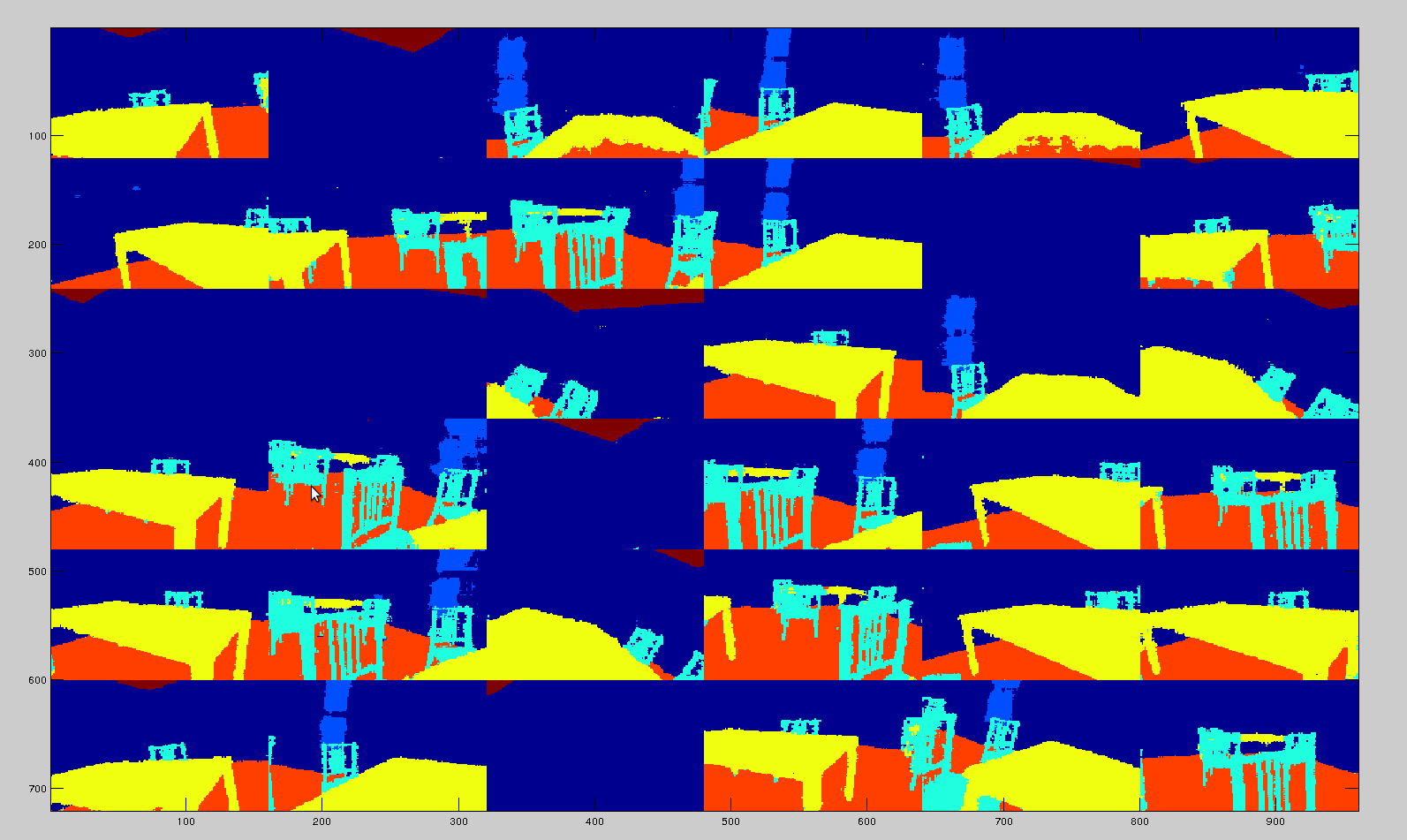}
}
\caption{\normalsize
Preliminary results of our architecture demonstrate the capabilities to jointly 
learn pixel-wise classifiers to produce a smooth segmentation. From top to 
bottom, we see how the four different layers of our architecture progressively 
improve the labels. Note that these results are on training set and the colour coding of labels is different.}
\label{fig:layers}
\end{figure*}

\vspace{-4mm}

%

\section{Conclusion}
We are working towards a real-time system for semantic scene understanding that 
combines the strengths of 3D reconstruction and semantic segmentation. 
We investigate the possibilities of using only depth data for this task and we make publicly available a new library containing the data and the code necessary to generate high-quality annotations for indoor scenes. Future work includes expanding the repository with new synthesised scenes \cite{Ruiqi:arXiv2015} to learn effective models for indoor semantic segmentation.	

\end{document}